%% file: main.tex
\documentclass{article}

\usepackage[final]{nips_2016}

\usepackage{tikz}
\usetikzlibrary{shapes, arrows, calc, positioning,matrix}
\tikzset{
data/.style={circle, draw, text centered, minimum height=3em ,minimum width = .5em, inner sep = 2pt},
empty/.style={circle, text centered, minimum height=3em ,minimum width = .5em, inner sep = 2pt},
}

\usepackage[utf8]{inputenc} 
\usepackage[T1]{fontenc}    
\usepackage{hyperref}       
\usepackage{url}            
\usepackage{booktabs}       
\usepackage{amsfonts}       
\usepackage{nicefrac}       
\usepackage{microtype}      
\usepackage{tikz}
\usetikzlibrary{positioning}
\usetikzlibrary{arrows}
\usepackage{dsfont}

\usepackage{times}
\usepackage{graphicx} 
\usepackage{caption}
\usepackage{wrapfig}

\usepackage{algorithm}
\usepackage{algorithmic}

\usepackage{hyperref}

\usepackage{amsmath}
\usepackage{amssymb}

\usepackage[font=footnotesize]{caption,subcaption}
\usepackage{sidecap}

\usepackage{xspace}

\usepackage{color}

\newcommand{\pushright}[1]{\ifmeasuring@#1\else\omit\hfill$\displaystyle#1$\fi\ignorespaces}

\include{defs}

\title{Reinforcement Learning and Control as Probabilistic Inference: Tutorial and Review}

\author{
  Sergey Levine \\
  UC Berkeley\\
  \texttt{svlevine@eecs.berkeley.edu} \\
}

\begin{document}

\maketitle


\begin{abstract}
The framework of reinforcement learning or optimal control provides a mathematical formalization of intelligent decision making that is powerful and broadly applicable. While the general form of the reinforcement learning problem enables effective reasoning about uncertainty, the connection between reinforcement learning and inference in probabilistic models is not immediately obvious. However, such a connection has considerable value when it comes to algorithm design: formalizing a problem as probabilistic inference in principle allows us to bring to bear a wide array of approximate inference tools, extend the model in flexible and powerful ways, and reason about compositionality and partial observability. In this article, we will discuss how a generalization of the reinforcement learning or optimal control problem, which is sometimes termed maximum entropy reinforcement learning, is equivalent to exact probabilistic inference in the case of deterministic dynamics, and variational inference in the case of stochastic dynamics. We will present a detailed derivation of this framework, overview prior work that has drawn on this and related ideas to propose new reinforcement learning and control algorithms, and describe perspectives on future research.
\end{abstract} 


\section{Introduction}

Probabilistic graphical models (PGMs) offer a broadly applicable and useful toolbox for the machine learning researcher~\citep{koller_friedman}: by couching the entirety of the learning problem in the parlance of probability theory, they provide a consistent and flexible framework to devise principled objectives, set up models that reflect the causal structure in the world, and allow a common set of inference methods to be deployed against a broad range of problem domains. Indeed, if a particular learning problem can be set up as a probabilistic graphical model, this can often serve as the first and most important step to solving it. Crucially, in the framework of PGMs, it is sufficient to write down the model and pose the question, and the objectives for learning and inference emerge automatically.

Conventionally, decision making problems formalized as reinforcement learning or optimal control have been cast into a framework that aims to generalize probabilistic models by augmenting them with utilities or rewards, where the reward function is viewed as an extrinsic signal. In this view, determining an optimal course of action (a plan) or an optimal decision-making strategy (a policy) is a fundamentally distinct type of problem than probabilistic inference, although the underlying dynamical system might still be described by a probabilistic graphical model. In this article, we instead derive an alterate view of decision making, reinforcement learning, and optimal control, where the decision making problem is simply an inference problem in a particular type of graphical model. Formalizing decision making as inference in probabilistic graphical models can in principle allow us to to bring to bear a wide array of approximate inference tools, extend the model in flexible and powerful ways, and reason about compositionality and partial observability.

Specifically, we will discuss how a generalization of the reinforcement learning or optimal control problem, which is sometimes termed maximum entropy reinforcement learning, is equivalent to exact probabilistic inference in the case of deterministic dynamics, and variational inference in the case of stochastic dynamics. This observation is not a new one, and the connection between probabilistic inference and control has been explored in the literature under a variety of names, including the Kalman duality~\citep{todorov_kalman_duality}, maximum entropy reinforcement learning~\citep{ziebart_thesis}, KL-divergence control~\citep{kappen_pgm,kappen_oc}, and stochastic optimal control~\citep{toussaint_soc}. While the specific derivations the differ, the basic underlying framework and optimization objective are the same. All of these methods involve formulating control or reinforcement learning as a PGM, either explicitly or implicitly, and then deploying learning and inference methods from the PGM literature to solve the resulting inference and learning problems.

Formulating reinforcement learning and decision making as inference provides a number of other appealing tools: a natural exploration strategy based on entropy maximization, effective tools for inverse reinforcement learning, and the ability to deploy powerful approximate inference algorithms to solve reinforcement learning problems. Furthermore, the connection between probabilistic inference and control provides an appealing probabilistic interpretation for the meaning of the reward function, and its effect on the optimal policy. The design of reward or cost functions in reinforcement learning is oftentimes as much art as science, and the choice of reward often blurs the line between algorithm and objective, with task-specific heuristics and task objectives combined into a single reward. In the control as inference framework, the reward induces a distribution over random variables, and the optimal policy aims to explicitly match a probability distribution defined by the reward and system dynamics, which may in future work suggest a way to systematize reward design.

This article will present the probabilistic model that can be used to embed a maximum entropy generalization of control or reinforcement learning into the framework of PGMs, describe how to perform inference in this model -- exactly in the case of deterministic dynamics, or via structured variational inference in the case of stochastic dynamics, -- and discuss how approximate methods based on function approximation fit within this framework. Although the particular variational inference interpretation of control differs somewhat from the presentation in prior work, the goal of this article is not to propose a fundamentally novel way of viewing the connection between control and inference. Rather, it is to provide a unified treatment of the topic in a self-contained and accessible tutorial format, and to connect this framework to recent research in reinforcement learning, including recently proposed deep reinforcement learning algorithms. In addition, this article presents a review of the recent reinforcement learning literature that relates to this view of control as probabilistic inference, and offers some perspectives on future research directions.

The basic graphical model for control will be presented in Section~\ref{sec:graphical_model}, variational inference for stochastic dynamics will be discussed in Section~\ref{sec:variational}, approximate methods based on function approximation, including deep reinforcement learning, will be discussed in Section~\ref{sec:deeprl}, and a survey and review of recent literature will be presented in Section~\ref{sec:related}. Finally, we will discuss perspectives on future research directions in Section~\ref{sec:future}.

\section{A Graphical Model for Control as Inference}
\label{sec:graphical_model}

In this section, we will present the basic graphical model that allows us to embed control into the framework of PGMs, and discuss how this framework can be used to derive variants of several standard reinforcement learning and dynamic programming approaches. The PGM presented in this section corresponds to a generalization of the standard reinforcement learning problem, where the RL objective is augmented with an entropy term. The magnitude of the reward function trades off between reward maximization and entropy maximization, allowing the original RL problem to be recovered in the limit of infinitely large rewards. We will begin by defining notation, then defining the graphical model, and then presenting several inference methods and describing how they relate to standard algorithms in reinforcement learning and dynamic programming. Finally, we will discuss a few limitations of this method and motivate the variational approach in Section~\ref{sec:variational}.

\subsection{The Decision Making Problem and Terminology}

First, we will introduce the notation we will use for the standard optimal control or reinforcement learning formulation. We will use $\bs \in \mathcal{S}$ to denote states and $\ba \in \mathcal{A}$ to denote actions, which may each be discrete or continuous. States evolve according to the stochastic dynamics $p(\bs_{t+1}|\bs_t,\ba_t)$, which are in general unknown. We will follow a discrete-time finite-horizon derivation, with horizon $T$, and omit discount factors for now. A discount $\gamma$ can be readily incorporated into this framework simply by modifying the transition dynamics, such that any action produces a transition into an absorbing state with probability $1-\gamma$, and all standard transition probabilities are multiplied by $\gamma$. A task in this framework can be defined by a reward function $r(\bs_t,\ba_t)$. Solving a task typically involves recovering a policy $p(\ba_t|\bs_t, \theta)$, which specifies a distribution over actions conditioned on the state parameterized by some parameter vector $\theta$. A standard reinforcement learning policy search problem is then given by the following maximization:
\begin{equation}
\theta^\star = \arg\max_\theta \sum_{t=1}^T E_{(\bs_t,\ba_t) \sim p(\bs_t,\ba_t|\theta)}[r(\bs_t,\ba_t)]. \label{eq:rl}
\end{equation}
This optimization problem aims to find a vector of policy parameters $\theta$ that maximize the total expected reward $\sum_t r(\bs_t,\ba_t)$ of the policy. The expectation is taken under the policy's \emph{trajectory} distribution $p(\tau)$, given by
\begin{equation}
p(\tau) = p(\bs_1,\ba_t,\dots,\bs_T,\ba_T|\theta) = p(\bs_1)\prod_{t=1}^T p(\ba_t|\bs_t, \theta) p(\bs_{t+1}|\bs_t,\ba_t).
\end{equation}
For conciseness, it is common to denote the action conditional $p(\ba_t|\bs_t, \theta)$ as $\pi_\theta(\ba_t|\bs_t)$, to emphasize that it is given by a parameterized policy with parameters $\theta$. These parameters might correspond, for example, to the weights in a neural network. However, we could just as well embed a standard planning problem in this formulation, by letting $\theta$ denote a sequence of actions in an open-loop plan.

Having formulated the decision making problem in this way, the next question we have to ask to derive the control as inference framework is: how can we formulate a probabilistic graphical model such that the most probable trajectory corresponds to the trajectory from the optimal policy? Or, equivalently, how can we formulate a probabilistic graphical model such that inferring the posterior action conditional $p(\ba_t|\bs_t, \theta)$ gives us the optimal policy?

\subsection{The Graphical Model}

\begin{figure}
    \centering
    
\tikzstyle{obs}=[shape=circle,draw=black!50,fill=gray!20]
\tikzstyle{state}=[shape=circle,draw=black!50,fill=white!20]
\tikzstyle{action}=[shape=circle,draw=black!50,fill=white!20]
\tikzstyle{lightedge}=[<-,thick]
\tikzstyle{bendedge}=[<-,bend left=45]

\begin{subfigure}{.5\textwidth}
  \centering

\begin{tikzpicture}[]

\node[state] (a1) at (0,2) {$\ba_1$};
\node[state] (a2) at (1,2) {$\ba_2$};
\node[state] (a3) at (2,2) {$\ba_3$};
\node[state] (a4) at (3,2) {$\ba_4$};

\node[action] (s1) at (0,1) {$\bs_1$};
\node[action] (s2) at (1,1) {$\bs_2$}
    edge [lightedge] (s1)
    edge [lightedge] (a1);
\node[action] (s3) at (2,1) {$\bs_3$}
    edge [lightedge] (s2)
    edge [lightedge] (a2);
\node[action] (s4) at (3,1) {$\bs_4$}
    edge [lightedge] (s3)
    edge [lightedge] (a3);

\end{tikzpicture}

  \caption{graphical model with states and actions}
  \label{fig:sub1}
\end{subfigure}%
\begin{subfigure}{.5\textwidth}
  \centering

\begin{tikzpicture}[]

\node[state] (a1) at (0,2) {$\ba_1$};
\node[state] (a2) at (1,2) {$\ba_2$};
\node[state] (a3) at (2,2) {$\ba_3$};
\node[state] (a4) at (3,2) {$\ba_4$};

\node[action] (s1) at (0,1) {$\bs_1$};
\node[action] (s2) at (1,1) {$\bs_2$}
    edge [lightedge] (s1)
    edge [lightedge] (a1);
\node[action] (s3) at (2,1) {$\bs_3$}
    edge [lightedge] (s2)
    edge [lightedge] (a2);
\node[action] (s4) at (3,1) {$\bs_4$}
    edge [lightedge] (s3)
    edge [lightedge] (a3);

\node[obs] (o1) at (0,3) {$\op_1$}
    edge [bendedge] (s1)
    edge [lightedge] (a1);
\node[obs] (o2) at (1,3) {$\op_2$}
    edge [bendedge] (s2)
    edge [lightedge] (a2);
\node[obs] (o3) at (2,3) {$\op_3$}
    edge [bendedge] (s3)
    edge [lightedge] (a3);
\node[obs] (o4) at (3,3) {$\op_4$}
    edge [bendedge] (s4)
    edge [lightedge] (a4);

\end{tikzpicture}

  \caption{graphical model with optimality variables}
  \label{fig:sub2}
\end{subfigure}

    \caption{The graphical model for control as inference. We begin by laying out the states and actions, which form the backbone of the model (a). In order to embed a control problem into this model, we need to add nodes that depend on the reward (b). These ``optimality variables'' correspond to observations in a HMM-style framework: we condition on the optimality variables being true, and then infer the most probable action sequence or most probable action distributions.}
    \label{fig:pgm_simple}
\end{figure}
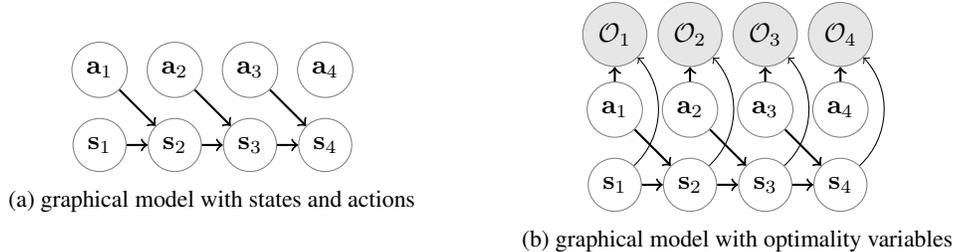

To embed the control problem into a graphical model, we can begin simply by modeling the relationship between states, actions, and next states. This relationship is simple, and corresponds to a graphical model with factors of the form $p(\bs_{t+1}|\bs_t,\ba_t)$, as shown in Figure~\ref{fig:pgm_simple} (a). However, this graphical model is insufficient for solving control problems, because it has no notion of rewards or costs. We therefore have to introduce an additional variable into this model, which we will denote $\op_t$. This additional variable is a binary random variable, where $\op_t = 1$ denotes that time step $t$ is \emph{optimal}, and $\op_t = 0$ denotes that it is not optimal. We will choose the distribution over this variable to be given by the following equation:
\begin{equation}
p(\op_t = 1 | \bs_t, \ba_t) = \exp(r(\bs_t,\ba_t)).\label{eq:opdef}
\end{equation}
The graphical model with these additional variables is summarized in Figure~\ref{fig:pgm_simple} (b). While this might at first seem like a peculiar and arbitrary choice, it leads to a very natural posterior distribution over actions when we condition on $\op_t = 1$ for all $t \in \{1, \dots, T\}$:
\begin{align}
p(\tau | \bo_{1:T}) \propto p(\tau, \bo_{1:T}) &= p(\bs_1)\prod_{t=1}^T p(\op_t = 1 | \bs_t, \ba_t) p(\bs_{t+1}|\bs_t,\ba_t) \nonumber \\
&= p(\bs_1)\prod_{t=1}^T \exp(r(\bs_t,\ba_t)) p(\bs_{t+1}|\bs_t,\ba_t) \nonumber \\
&= \left[ p(\bs_1) \prod_{t=1}^T p(\bs_{t+1}|\bs_t,\ba_t) \right] \exp\left( \sum_{t=1}^T r(\bs_t,\ba_t) \right).\label{eq:graphicalmodel}
\end{align}
That is, the probability of observing a given trajectory is given by the product between its probability to occur according to the dynamics (the term in square brackets on the last line), and the exponential of the total reward along that trajectory. It is most straightforward to understand this equation in systems with deterministic dynamics, where the first term is a constant for all trajectories that are dynamically feasible. In this case, the trajectory with the highest reward has the highest probability, and trajectories with lower reward have exponentially lower probability. If we would like to plan for an optimal action sequence starting from some initial state $\bs_1$, we can condition on $\bo_{1:T}$ and choose $p(\bs_1) = \delta(\bs_1)$, in which case maximum a posteriori inference corresponds to a kind of planning problem. It is easy to see that this exactly corresponds to standard planning or trajectory optimization in the case where the dynamics are deterministic, in which case Equation~(\ref{eq:graphicalmodel}) reduces to
\begin{equation}
p(\tau | \bo_{1:T}) \propto \mathds{1}[p(\tau) \neq 0] \exp\left( \sum_{t=1}^T r(\bs_t,\ba_t) \right). \label{eq:detmaxent}
\end{equation}
Here, the indicator function simply indicates that the trajectory $\tau$ is dynamically consistent (meaning that $p(\bs_{t+1} | \bs_t,\ba_t) \neq 0$) and the initial state is correct. The case of stochastic dynamics poses some challenges, and will be discussed in detail in Section~\ref{sec:variational}. However, even under deterministic dynamics, we are often interested in recovering a policy rather than a plan. In this PGM, the optimal policy can be written as $p(\ba_t|\bs_t,\op_{t:T} = 1)$ (we will drop $=1$ in the remainder of the derivation for conciseness). This distribution is somewhat analogous to $p(\ba_t|\bs_t, \theta^\star)$ in the previous section, with two major differences: first, it is independent of the parameterization $\theta$, and second, we will see later that it optimizes an objective that is slightly different from the standard reinforcement learning objective in Equation~(\ref{eq:rl}).

\subsection{Policy Search as Probabilistic Inference}
\label{sec:maxentinference}

We can recover the optimal policy $p(\ba_t|\bs_t,\op_{t:T})$ using a standard sum-product inference algorithm, analogously to inference in HMM-style dynamic Bayesian networks. As we will see in this section, it is sufficient to compute backward messages of the form
\begin{equation}
\beta_t(\bs_t,\ba_t) = p(\op_{t:T}|\bs_t,\ba_t). \nonumber
\end{equation}
These messages have a natural interpretation: they denote the probability that a trajectory can be optimal for time steps from $t$ to $T$ if it begins in state $\bs_t$ with the action $\ba_t$.\footnote{Note that $\beta_t(\bs_t,\ba_t)$ is \emph{not} a probability density over $\bs_t, \ba_t$, but rather the probability of $\op_{t:T} = 1$.} Slightly overloading the notation, we will also introduce the message
\begin{equation}
\beta_t(\bs_t) = p(\op_{t:T}|\bs_t). \nonumber
\end{equation}
These messages denote the probability that the trajectory from $t$ to $T$ is optimal if it begins in state $\bs_t$. We can recover the state-only message from the state-action message by integrating out the action:
\begin{align}
\beta_t(\bs_t) = p(\op_{t:T}|\bs_t) = \int_{\mathcal{A}} p(\op_{t:T}|\bs_t,\ba_t) p(\ba_t|\bs_t) d\ba_t = \int_{\mathcal{A}} \beta_t(\bs_t,\ba_t) p(\ba_t|\bs_t) d\ba_t. \nonumber
\end{align}
The factor $p(\ba_t|\bs_t)$ is the action \emph{prior}. Note that it is not conditioned on $\op_{1:T}$ in any way: it does not denote the probability of an optimal action, but simply the prior probability of actions. The PGM in Figure~\ref{fig:pgm_simple} doesn't actually contain this factor, and we can assume that $p(\ba_t|\bs_t) = \frac{1}{|\mathcal{A}|}$ for simplicity -- that is, it is a constant corresponding to a uniform distribution over the set of actions. We will see later that this assumption does not actually introduce any loss of generality, because any non-uniform $p(\ba_t|\bs_t)$ can be incorporated instead into $p(\op_t | \bs_t, \ba_t)$ via the reward function.

The recursive message passing algorithm for computing $\beta_t(\bs_t,\ba_t)$ proceeds from the last time step $t=T$ backward through time to $t=1$. In the base case, we note that $p(\op_T|\bs_T,\ba_T)$ is simply proportional to $\exp(r(\bs_T,\ba_T))$, since there is only one factor to consider. The recursive case is then given as following:
\begin{align}
\beta_t(\bs_t,\ba_t) = p(\op_{t:T}|\bs_t,\ba_t) = \int_{\mathcal{S}} \beta_{t+1}(\bs_{t+1}) p(\bs_{t+1}|\bs_t,\ba_t) p(\op_t|\bs_t,\ba_t) d\bs_{t+1}.\label{eq:betabackup}
\end{align}
From these backward messages, we can then derive the optimal policy $p(\ba_t|\bs_t,\op_{1:T})$. First, note that $\op_{1:(t-1)}$ is conditionally independent of $\ba_t$ given $\bs_t$, which means that $p(\ba_t|\bs_t,\op_{1:T})=p(\ba_t|\bs_t,\op_{t:T})$, and we can disregard the past when considering the current action distribution. This makes intuitive sense: in a Markovian system, the optimal action does not depend on the past. From this, we can easily recover the optimal action distribution using the two backward messages:
\begin{align}
p(\ba_t|\bs_t,\op_{t:T}) = \frac{p(\bs_t,\ba_t|\op_{t:T})}{p(\bs_t|\op_{t:T})} =
\frac{p(\op_{t:T}|\bs_t,\ba_t)p(\ba_t|\bs_t)p(\bs_t)}{p(\op_{t:T}|\bs_t)p(\bs_t)} \propto
\frac{p(\op_{t:T}|\bs_t,\ba_t)}{p(\op_{t:T}|\bs_t)} =
\frac{\beta_t(\bs_t,\ba_t)}{\beta_t(\bs_t)}, \nonumber
\end{align}
where the order of conditioning in the third step is flipped by using Bayes' rule, and cancelling the factor of $p(\op_{t:T})$ that appears in both the numerator and denominator. The term $p(\ba_t|\bs_t)$ disappears, since we previously assumed it was a uniform distribution.

This derivation provides us with a solution, but perhaps not as much of the intuition. The intuition can be recovered by considering what these equations are doing in log space. To that end, we will introduce the log-space messages as
\begin{align*}
Q(\bs_t,\ba_t) &= \log \beta_t(\bs_t,\ba_t) \\
V(\bs_t) &= \log \beta_t(\bs_t).
\end{align*}
The use of $Q$ and $V$ here is not accidental: the log-space messages correspond to ``soft'' variants of the state and state-action value functions. First, consider the marginalization over actions in log-space:
\begin{equation*}
V(\bs_t) = \log \int_\mathcal{A} \exp(Q(\bs_t,\ba_t)) d\ba_t.
\end{equation*}
When the values of $Q(\bs_t,\ba_t)$ are large, the above equation resembles a hard maximum over $\ba_t$. That is, for large $Q(\bs_t,\ba_t)$,
\begin{equation*}
V(\bs_t) = \log \int_\mathcal{A} \exp(Q(\bs_t,\ba_t)) d\ba_t \approx \max_{\ba_t} Q(\bs_t,\ba_t).
\end{equation*}
For smaller values of $Q(\bs_t,\ba_t)$, the maximum is soft. Hence, we can refer to $V$ and $Q$ as soft value functions and Q-functions, respectively. We can also consider the backup in Equation~(\ref{eq:betabackup}) in log-space. In the case of deterministic dynamics, this backup is given by
\begin{equation*}
Q(\bs_t,\ba_t) = r(\bs_t,\ba_t) + V(\bs_{t+1}),
\end{equation*}
which exactly corresponds to the Bellman backup. However, when the dynamics are stochastic, the backup is given by
\begin{equation}
Q(\bs_t,\ba_t) = r(\bs_t,\ba_t) + \log E_{\bs_{t+1} \sim p(\bs_{t+1}|\bs_t,\ba_t)} [\exp( V(\bs_{t+1}))].\label{eq:optimisticbackup}
\end{equation}
This backup is peculiar, since it does not consider the expected value at the next state, but a ``soft max'' over the next expected value. Intuitively, this produces Q-functions that are optimistic: if among the possible outcomes for the next state there is one outcome with a very high value, it will dominate the backup, even when there are other possible states that might be likely and have extremely low value. This creates risk seeking behavior: if an agent behaves according to this Q-function, it might take actions that have extremely high risk, so long as they have some non-zero probability of a high reward. Clearly, this behavior is not desirable in many cases, and the standard PGM described in this section is often not well suited to stochastic dynamics. In Section~\ref{sec:variational}, we will describe a simple modification that makes the backup correspond to the soft Bellman backup in the case of stochastic dynamics also, by using the framework of variational inference.

\subsection{Which Objective does This Inference Procedure Optimize?}
\label{sec:maxentobjective}

In the previous section, we derived an inference procedure that can be used to obtain the distribution over actions conditioned on all of the optimality variables, $p(\ba_t|\bs_t,\op_{1:T})$. But which objective does this policy actually optimize? Recall that the overall distribution is given by
\begin{align}
p(\tau) = \left[ p(\bs_1) \prod_{t=1}^T p(\bs_{t+1}|\bs_t,\ba_t) \right] \exp\left( \sum_{t=1}^T r(\bs_t,\ba_t) \right), \label{eq:maxentdistro}
\end{align}
which we can simplify in the case of deterministic dynamics into Equation~(\ref{eq:detmaxent}). In this case, the conditional distributions $p(\ba_t|\bs_t,\op_{1:T})$ are simply obtained by marginalizing the full trajectory distribution and conditioning the policy at each time step on $\bs_t$. We can adopt an optimization-based approximate inference approach to this problem, in which case the goal is to fit an approximation $\pi(\ba_t|\bs_t)$ such that the trajectory distribution
\begin{align*}
\hat{p}(\tau) \propto \mathds{1}[p(\tau) \neq 0] \prod_{t=1}^T \pi(\ba_t|\bs_t)
\end{align*}
matches the distribution in Equation~(\ref{eq:detmaxent}). In the case of exact inference, as derived in the previous section, the match is exact, which means that $\kl(\hat{p}(\tau) \| p(\tau)) = 0$, where $\kl$ is the KL-divergence. We can therefore view the inference process as minimizing $\kl(\hat{p}(\tau) \| p(\tau))$, which is given by
\begin{equation*}
\kl(\hat{p}(\tau) \| p(\tau)) = -E_{\tau \sim \hat{p}(\tau)}[\log p(\tau) - \log \hat{p}(\tau)].
\end{equation*}
Negating both sides and substituting in the equations for $p(\tau)$ and $\hat{p}(\tau)$, we get
\begin{align*}
-\kl(\hat{p}(\tau) \| p(\tau)) &= E_{\tau \sim \hat{p}(\tau)}\left[ \log p(\bs_1) +  \sum_{t=1}^T \left( \log p(\bs_{t+1} |\bs_t,\ba_t) + r(\bs_t,\ba_t) \right) - \right. \\
& \hspace{0.73in} \left. \log p(\bs_1) - \sum_{t=1}^T \left( \log p(\bs_{t+1} | \bs_t,\ba_t) + \log\pi(\ba_t|\bs_t) \right) \right] \\
&= E_{\tau \sim \hat{p}(\tau)}\left[ \sum_{t=1}^T r(\bs_t,\ba_t) - \log\pi(\ba_t|\bs_t) \right] \\
&= \sum_{t=1}^T E_{(\bs_t,\ba_t) \sim \hat{p}(\bs_t,\ba_t))}[r(\bs_t,\ba_t) - \log\pi(\ba_t|\bs_t)] \\
&= \sum_{t=1}^T E_{(\bs_t,\ba_t) \sim \hat{p}(\bs_t,\ba_t))}[r(\bs_t,\ba_t)] + E_{\bs_t \sim \hat{p}(\bs_t)}[ \ent(\pi(\ba_t|\bs_t))].\label{eq:maxentpolicy}
\end{align*}
Therefore, minimizing the KL-divergence corresponds to maximizing the expected reward \emph{and} the expected conditional entropy, in contrast to the standard control objective in Equation~(\ref{eq:rl}), which only maximizes reward. Hence, this type of control objective is sometimes referred to as maximum entropy reinforcement learning or maximum entropy control.

However, that in the case of stochastic dynamics, the solution is not quite so simple. Under stochastic dynamics, the optimized distribution is given by
\begin{align}
\hat{p}(\tau) = p(\bs_1 | \op_{1:T}) \prod_{t=1}^T p(\bs_{t+1}|\bs_t,\ba_t, \op_{1:T}) p(\ba_t|\bs_t, \op_{1:T}),
\end{align}
where the initial state distribution and the dynamics are \emph{also} conditioned on optimality. As a result of this, the dynamics and initial state terms in the KL-divergence do \emph{not} cancel, and the objective does not have the simple entropy maximizing form derived above.\footnote{In the deterministic case, we know that $p(\bs_{t+1}|\bs_t,\ba_t, \op_{1:T}) = p(\bs_{t+1}|\bs_t,\ba_t)$, since exactly one transition is ever possible.} We can still fall back on the original KL-divergence minimization at the trajectory level, and write the objective as
\begin{equation}
-\kl(\hat{p}(\tau) \| p(\tau)) = E_{\tau \sim \hat{p}(\tau)}\left[ \log p(\bs_1) + \sum_{t=1}^T r(\bs_t,\ba_t) + \log p(\bs_{t+1} |\bs_t,\ba_t) \right] + \ent(\hat{p}(\tau)). \label{eq:maxentstochdynamics}
\end{equation}
However, because of the $\log p(\bs_{t+1} |\bs_t,\ba_t)$ terms, this objective is difficult to optimize in a model-free setting. As discussed in the previous section, it also results in an optimistic policy that assumes a degree of control over the dynamics that is unrealistic in most control problems. In Section~\ref{sec:variational}, we will derive a variational inference procedure that \emph{does} reduce to the convenient objective in Equation~(\ref{eq:maxentpolicy}) even in the case of stochastic dynamics, and in the process also addresses the risk-seeking behavior discussed in Section~\ref{sec:maxentinference}.

\subsection{Alternative Model Formulations}
\label{sec:alt_model}

It's worth pointing out that the definition of $p(\op_t = 1 | \bs_t, \ba_t)$ in Equation~(\ref{eq:opdef}) requires an additional assumption, which is that the rewards $r(\bs_t,\ba_t)$ are always negative.\footnote{This assumption is not actually very strong: if we assume the reward is bounded above, we can always construct an exactly equivalent reward simply by subtracting the maximum reward.} Otherwise, we end up with a negative probability for $p(\op_t = 0 | \bs_t, \ba_t)$. However, this assumption is not actually required: it's quite possible to instead define the graphical model with an undirected factor on $(\bs_t,\ba_t,\op_t)$, with an unnormalized potential given by $\Phi_t(\bs_t,\ba_t,\op_t) = \mathds{1}_{\op_t = 1}\exp(r(\bs_t,\ba_t))$. The potential for $\op_t = 0$ doesn't matter, since we always condition on $\op_t = 1$. This leads to the same exact inference procedure as the one we described above, but without the negative reward assumption. Once we are content to working with undirected graphical models, we can even remove the variables $\op_t$ completely, and simply add an undirected factor on $(\bs_t,\ba_t)$ with the potential $\Phi_t(\bs_t,\ba_t) = \exp(r(\bs_t,\ba_t))$, which is mathematically equivalent. This is the conditional random field formulation described by Ziebart~\citep{ziebart_thesis}. The analysis and inference methods in this model are identical to the ones for the directed model with explicit optimality variables $\op_t$, and the particular choice of model is simply a notational convenience. We will use the variables $\op_t$ in this article for clarity of derivation and stay within the directed graphical model framework, but all derivations are straightforward to reproduce in the conditional random field formulation.

Another common modification to this framework is to incorporate an explicit temperature $\alpha$ into the CPD for $\opt$, such that $p(\opt|\bs_t,ba_t) = \exp(\frac{1}{\alpha} r(\bs_t,\ba_t))$. The corresponding maximum entropy objective can then be written equivalently as the expectation of the (original) reward, with an additional multiplier of $\alpha$ on the entropy term. This provides a natural mechanism to interpolate between entropy maximization and standard optimal control or RL: as $\alpha \rightarrow 0$, the optimal solution approaches the standard optimal control solution. Note that this does not actually increase the generality of the method, since the constant $\frac{1}{\alpha}$ can always be multiplied into the reward, but making this temperature constant explicit can help to illuminate the connection between standard and entropy maximizing optimal control.

Finally, it is worth remarking again on the role of discount factors: it is very common in reinforcement learning to use a Bellman backup of the form
\[
Q(\bs_t,\ba_t) \leftarrow r(\bs_t,\ba_t) + \gamma E_{\bs_{t+1} \sim p(\bs_{t+1}|\bs_t,\ba_t)}[V(\bs_{t+1})],
\]
where $\gamma \in (0,1]$ is a discount factor. This allows for learning value functions in infinite-horizon settings, where the backup would otherwise be non-convergent for $\gamma = 1$, and reduces variance for Monte Carlo advantage estimators in policy gradient algorithms~\citep{gae}. The discount factor can be viewed a simple redefinition of the system dynamics. If the initial dynamics are given by $p(\bs_{t+1}|\bs_t,\ba_t)$, adding a discount factor is equivalent to undiscounted value fitting under the modified dynamics $\bar{p}(\bs_{t+1}|\bs_t,\ba_t) = \gamma p(\bs_{t+1}|\bs_t,\ba_t)$, where there is an additional transition with probability $1-\gamma$, regardless of action, into an absorbing state with reward zero. We will omit $\gamma$ from the derivations in this article, but it can be inserted trivially in all cases simply by modifying the (soft) Bellman backups in any place where the expectation over $p(\bs_{t+1}|\bs_t,\ba_t)$ occurs, such as Equation~(\ref{eq:optimisticbackup}) previously or Equation~(\ref{eq:bellmanbackup}) in the next section.

\section{Variational Inference and Stochastic Dynamics}
\label{sec:variational}

The problematic nature of the maximum entropy framework in the case of stochastic dynamics, discussed in Section~\ref{sec:maxentinference} and Section~\ref{sec:maxentobjective}, in essence amounts to an assumption that the agent is allowed to control both its actions and the dynamics of the system in order to produce optimal trajectories, but its authority over the dynamics is penalized based on deviation from the true dynamics. Hence, the $\log p(\bs_{t+1} |\bs_t,\ba_t)$ terms in Equation~(\ref{eq:maxentstochdynamics}) can be factored out of the equations, producing additive terms that corresponds to the cross-entropy between the posterior dynamics $p(\bs_{t+1}|\bs_t,\ba_t, \op_{1:T})$ and the true dynamics $p(\bs_{t+1} |\bs_t,\ba_t)$. This explains the risk-seeking nature of the method discussed in Section~\ref{sec:maxentinference}: if the agent is allowed to influence its dynamics, even a little bit, it would reasonably choose to remove unlikely but extremely bad outcomes of risky actions.

Of course, in practical reinforcement learning and control problems, such manipulation of system dynamics is not possible, and the resulting policies can lead to disastrously bad outcomes. We can correct this issue by modifying the inference procedure. In this section, we will derive this correction by freezing the system dynamics, writing down the corresponding maximum entropy objective, and deriving a dynamic programming procedure for optimizing it. Then we will show that this procedure amounts to a direct application of structured variational inference.

\subsection{Maximum Entropy Reinforcement Learning with Fixed Dynamics}
\label{sec:fixeddynamics}

The issue discussed in Section~\ref{sec:maxentobjective} for stochastic dynamics can briefly be summarized as following: since the posterior dynamics distribution $p(\bs_{t+1}|\bs_t,\ba_t, \op_{1:T})$ does not necessarily match the true dynamics $p(\bs_{t+1}|\bs_t,\ba_t)$, the agent assumes that it can influence the dynamics to a limited extent. A simple fix to this issue is to explicitly disallow this control, by forcing the posterior dynamics and initial state distributions to match $p(\bs_{t+1}|\bs_t,\ba_t)$ and $p(\bs_1)$, respectively. Then, the optimized trajectory distribution is given simply by
\begin{align*}
\hat{p}(\tau) = p(\bs_1) \prod_{t=1}^T p(\bs_{t+1}|\bs_t,\ba_t) \pi(\ba_t|\bs_t),
\end{align*}
and the same derivation as the one presented in Section~\ref{sec:maxentobjective} for the deterministic case results in the following objective:
\begin{align}
-\kl(\hat{p}(\tau) \| p(\tau)) &= \sum_{t=1}^T E_{(\bs_t,\ba_t) \sim \hat{p}(\bs_t,\ba_t))}[r(\bs_t,\ba_t) + \ent(\pi(\ba_t|\bs_t))].\label{eq:maxcausalentpolicy}
\end{align}
That is, the objective is still to maximize reward and entropy, but now under stochastic transition dynamics. To optimize this objective, we can compute backward messages like we did in Section~\ref{sec:maxentinference}. However, since we are now starting from the maximization of the objective in Equation~(\ref{eq:maxcausalentpolicy}), we have to derive these backward messages from an optimization perspective as a dynamic programming algorithm. As before, we will begin with the base case of optimizing $\pi(\ba_T|\bs_T)$, which maximizes
\begin{align}
&E_{(\bs_T,\ba_T)\sim \hat{p}(\bs_T,\ba_T)}[r(\bs_T,\ba_T) - \log \pi(\ba_T|\bs_T)] = \nonumber \\
&E_{\bs_T \sim \hat{p}(\bs_T)}\left[ - \kl\left( \pi(\ba_T|\bs_T) \| \frac{1}{\exp(V(\bs_T))}\exp(r(\bs_T,\ba_T)) \right) + V(\bs_T) \right],\label{eq:maxcausalentkl}
\end{align}
where the equality holds from the definition of KL-divergence, and $\exp(V(\bs_T))$ is the normalizing constant for $\exp(r(\bs_T,\ba_T))$ with respect to $\ba_T$ where $V(\bs_T) = \log \int_{\mathcal{A}} \exp(r(\bs_T,\ba_T)) d\ba_T$, which is the same soft maximization as in Section~\ref{sec:maxentinference}. Since we know that the KL-divergence is minimized when the two arguments represent the same distribution, the optimal policy is given by
\begin{equation}
\pi(\ba_T | \bs_T) = \exp \left(
r(\bs_T,\ba_T) - V(\bs_T)
\right), \label{eq:maxcausalentbasecase}
\end{equation}
The recursive case can then computed as following: for a given time step $t$, $\pi(\ba_t | \bs_t)$ must maximize two terms:
\begin{align}
&E_{(\bs_t,\ba_t)\sim \hat{p}(\bs_t,\ba_t)}[r(\bs_t,\ba_t) - \log \pi(\ba_t|\bs_t)] + E_{(\bs_t,\ba_t)\sim \hat{p}(\bs_t,\ba_t)}[ E_{\bs_{t+1} \sim p(\bs_{t+1}|\bs_t,\ba_t)} [V(\bs_{t+1})] ].\label{eq:maxcentreccase}
\end{align}
The first term follows directly from the objective in Equation~(\ref{eq:maxcausalentpolicy}), while the second term represents the contribution of $\pi(\ba_t|\bs_t)$ to the expectations of all subsequent time steps. The second term deserves a more in-depth derivation. First, consider the base case: given the equation for $\pi(\ba_T | \bs_T)$ in Equation~(\ref{eq:maxcausalentbasecase}), we can evaluate the objective for the policy by directly substituting this equation into Equation~(\ref{eq:maxcausalentkl}). Since the KL-divergence then evaluates to zero, we are left only with the $V(\bs_T)$ term. In the recursive case, we note that we can rewrite the objective in Equation~(\ref{eq:maxcentreccase}) as
\begin{align*}
&E_{(\bs_t,\ba_t)\sim \hat{p}(\bs_t,\ba_t)}[r(\bs_t,\ba_t) - \log \pi(\ba_t|\bs_t)] + E_{(\bs_t,\ba_t)\sim \hat{p}(\bs_t,\ba_t)}[ E_{\bs_{t+1} \sim p(\bs_{t+1}|\bs_t,\ba_t)} [V(\bs_{t+1})] ] = \\
&E_{\bs_t \sim \hat{p}(\bs_t)}\left[ - \kl\left( \pi(\ba_t|\bs_t) \| \frac{1}{\exp(V(\bs_t))}\exp(Q(\bs_t,\ba_t)) \right) + V(\bs_t) \right],
\end{align*}
where we now define
\begin{align}
Q(\bs_t,\ba_t) &= r(\bs_t,\ba_t) + E_{\bs_{t+1} \sim p(\bs_{t+1}|\bs_t,\ba_t)}[V(\bs_{t+1})] \label{eq:bellmanbackup}\\
V(\bs_t) &= \log \int_{\mathcal{A}} \exp(Q(\bs_t,\ba_t)) d\ba_t, \nonumber
\end{align}
which corresponds to a standard Bellman backup with a soft maximization for the value function. Choosing
\begin{equation}
\pi(\ba_t | \bs_t) = \exp \left(
Q(\bs_t,\ba_t) - V(\bs_t)
\right), \label{eq:maxcausalentqpolicy}
\end{equation}
we again see that the KL-divergence evaluates to zero, leaving $E_{\bs_t \sim \hat{p}(\bs_t)}[V(\bs_t)]$ as the only remaining term in the objective for time step $t$, just like in the base case of $t=T$. This means that, if we fix the dynamics and initial state distribution, and only allow the policy to change, we recover a Bellman backup operator that uses the expected value of the next state, rather than the optimistic estimate we saw in Section~\ref{sec:maxentinference} (compare Equation~(\ref{eq:bellmanbackup}) to Equation~(\ref{eq:optimisticbackup})). While this provides a solution to the practical problem of risk-seeking policies, it is perhaps a bit unsatisfying in its divergence from the convenient framework of probabilistic graphical models. In the next section, we will discuss how this procedure amounts to a direct application of structured variational inference.

\subsection{Connection to Structured Variational Inference}

One way to interpret the optimization procedure in Section~\ref{sec:fixeddynamics} is as a particular type of structured variational inference. In structured variational inference, our goal is to approximate some distribution $p(\mathbf{y})$ with another, potentially simpler distribution $q(\mathbf{y})$. Typically, $q(\mathbf{y})$ is taken to be some tractable factorized distribution, such as a product of conditional distributions connected in a chain or tree, which lends itself to tractable exact inference. In our case, we aim to approximate $p(\tau)$, given by
\begin{align}
p(\tau) = \left[ p(\bs_1) \prod_{t=1}^T p(\bs_{t+1}|\bs_t,\ba_t) \right] \exp\left( \sum_{t=1}^T r(\bs_t,\ba_t) \right),\label{eq:variationaltrajdistrotarget}
\end{align}
via the distribution
\begin{align}
q(\tau) = q(\bs_1) \prod_{t=1}^T q(\bs_{t+1}|\bs_t,\ba_t) q(\ba_t|\bs_t).\label{eq:variationaltrajdistropol}
\end{align}
If we fix $q(\bs_1) = p(\bs_1)$ and $q(\bs_{t+1}|\bs_t,\ba_t) = p(\bs_{t+1}|\bs_t,\ba_t)$, then $q(\tau)$ is exactly the distribution $\hat{p}(\tau)$ from Section~\ref{sec:fixeddynamics}, which we've renamed here to $q(\tau)$ to emphasize the connection to structured variational inference. Note that we've also renamed $\pi(\ba_t|\bs_t)$ to $q(\ba_t|\bs_t)$ for the same reason. In structured variational inference, approximate inference is performed by optimizing the variational lower bound (also called the evidence lower bound). Recall that our evidence here is that $\op_t = 1$ for all $t \in \{1,\dots,T\}$, and the posterior is conditioned on the initial state $\bs_1$. The variational lower bound is given by
\begin{align*}
\log p(\op_{1:T}) &= \log \int \int p(\op_{1:T}, \bs_{1:T}, \ba_{1:T}) d\bs_{1:T} d\ba_{1:T} \\
&= \log \int \int p(\op_{1:T}, \bs_{1:T}, \ba_{1:T}) \frac{q(\bs_{1:T}, \ba_{1:T})}{q(\bs_{1:T}, \ba_{1:T})} d\bs_{1:T} d\ba_{1:T} \\
&= \log E_{(\bs_{1:T},\ba_{1:T})\sim q(\bs_{1:T}, \ba_{1:T})} \left[ \frac{p(\op_{1:T}, \bs_{1:T}, \ba_{1:T})}{q(\bs_{1:T}, \ba_{1:T})} \right] \\
&\geq E_{(\bs_{1:T},\ba_{1:T})\sim q(\bs_{1:T}, \ba_{1:T})} \left[ \log p(\op_{1:T}, \bs_{1:T}, \ba_{1:T}) - \log q(\bs_{1:T}, \ba_{1:T}) \right],
\end{align*}
where the inequality on the last line is obtained via Jensen's inequality. Substituting the definitions of $p(\tau)$ and $q(\tau)$ from Equations~(\ref{eq:variationaltrajdistrotarget}) and (\ref{eq:variationaltrajdistropol}), and noting the cancellation due to $q(\bs_{t+1}|\bs_t,\ba_t) = p(\bs_{t+1}|\bs_t,\ba_t)$, the bound reduces to
\begin{equation}
\log p(\op_{1:T}) \geq E_{(\bs_{1:T},\ba_{1:T})\sim q(\bs_{1:T}, \ba_{1:T})} \left[ \sum_{t=1}^T r(\bs_t,\ba_t) - \log q(\ba_t|\bs_t) \right],
\end{equation}
up to an additive constant. Optimizing this objective with respect to the policy $q(\ba_t|\bs_t)$ corresponds exactly to the objective in Equation~(\ref{eq:maxcausalentpolicy}). Intuitively, this means that this objective attempts to find the closest match to the maximum entropy trajectory distribution, subject to the constraint that the agent is only allowed to modify the policy, and not the dynamics. Note that this framework can also easily accommodate any other structural constraints on the policy, including restriction to a particular distribution class (e.g., conditional Gaussian, or a categorical distribution parameterized by a neural network), or restriction to partial observability, where the entire state $\bs_t$ is not available as an input, but rather the policy only has access to some non-invertible function of the state.

\section{Approximate Inference with Function Approximation}
\label{sec:deeprl}

We saw in the discussion above that a dynamic programming backward algorithm with updates that resemble Bellman backups can recover ``soft'' analogues of the value function and Q-function in the maximum entropy reinforcement learning framework, and the stochastic optimal policy can be recovered from the Q-function and value function. In this section, we will discuss how practical algorithms for high-dimensional or continuous reinforcement learning problems can be derived from this theoretical framework, with the use of function approximation. This will give rise to several prototypical methods that mirror corresponding techniques in standard reinforcement learning: policy gradients, actor-critic algorithms, and Q-learning.

\subsection{Maximum Entropy Policy Gradients}
\label{sec:maxentpg}

One approach to performing structured variational inference is to directly optimize the evidence lower bound with respect to the variational distribution~\citep{koller_friedman}. This approach can be directly applied to maximum entropy reinforcement learning. Note that the variational distribution consists of three terms: $q(\bs_1)$, $q(\bs_{t+1}|\bs_t,\ba_t)$, and $q(\ba_t|\bs_t)$. The first two terms are fixed to $p(\bs_1)$ and $p(\bs_{t+1}|\bs_t,\ba_t)$, respectively, leaving only $q(\ba_t|\bs_t)$ to vary. We can parameterize this distribution with any expressive conditional, with parameters $\theta$, and will therefore denote it as $q_\theta(\ba_t|\bs_t)$. The parameters could correspond, for example, to the weights in a deep neural network, which takes $\bs_t$ as input and outputs the parameters of some distribution class. In the case of discrete actions, the network could directly output the parameters of a categorical distribution (e.g., via a soft max operator). In the case of continuous actions, the network could output the parameters of an exponential family distribution, such as a Gaussian. In all cases, we can directly optimize the objective in Equation~(\ref{eq:maxcausalentpolicy}) by estimating its gradient using samples. This gradient has a form that is nearly identical to the standard policy gradient~\citep{williams}, which we summarize here for completeness. First, let us restate the objective as following:
\begin{align*}
J(\theta) = \sum_{t=1}^T E_{(\bs_t,\ba_t) \sim q(\bs_t,\ba_t)}\left[ r(\bs_t,\ba_t) - \mathcal{H}(q_\theta(\ba_t|\bs_t)) \right].
\end{align*}
The gradient is then given by
\begin{align*}
\nabla_\theta J(\theta) &= \sum_{t=1}^T \nabla_\theta E_{(\bs_t,\ba_t) \sim q(\bs_t,\ba_t)}\left[ r(\bs_t,\ba_t) + \mathcal{H}(q_\theta(\ba_t|\bs_t)) \right] \\
&= \sum_{t=1}^T E_{(\bs_t,\ba_t) \sim q(\bs_t,\ba_t)}\left[
\nabla_\theta \log q_\theta(\ba_t|\bs_t) \left(
\sum_{t'=t}^T r(\bs_{t'},\ba_{t'}) - \log q_\theta(\ba_{t'}|\bs_{t'}) - 1
\right)
\right] \\
&= \sum_{t=1}^T E_{(\bs_t,\ba_t) \sim q(\bs_t,\ba_t)}\left[
\nabla_\theta \log q_\theta(\ba_t|\bs_t) \left(
\sum_{t'=t}^T r(\bs_{t'},\ba_{t'}) - \log q_\theta(\ba_{t'}|\bs_{t'}) - b(\bs_{t'})
\right)
\right],
\end{align*}
where the second line follows from applying the likelihood ratio trick~\citep{williams} and the definition of entropy to obtain the $\log q_\theta(\ba_{t'}|\bs_{t'})$ term. The $-1$ comes from the derivative of the entropy term. The last line follows by noting that the gradient estimator is invariant to additive state-dependent constants, and replacing $-1$ with a state-dependent baseline $b(\bs_{t'})$. The resulting policy gradient estimator exactly matches a standard policy gradient estimator, with the only modification being the addition of the $-\log q_\theta(\ba_{t'}|\bs_{t'})$ term to the reward at each time step $t'$. Intuitively, the reward of each action is modified by subtracting the log-probability of that action under the current policy, which causes the policy to maximize entropy. This gradient estimator can be written more compactly as
\begin{align*}
\nabla_\theta J(\theta) &= \sum_{t=1}^T E_{(\bs_t,\ba_t) \sim q(\bs_t,\ba_t)}\left[
\nabla_\theta \log q_\theta(\ba_t|\bs_t) \hat{A}(\bs_t,\ba_t)
\right],
\end{align*}
where $\hat{A}(\bs_t,\ba_t)$ is an advantage estimator. Any standard advantage estimator, such as the GAE estimator~\citep{gae}, can be used in place of the standard baselined Monte Carlo return above. Again, the only necessary modification is to add $-\log q_\theta(\ba_{t'}|\bs_{t'})$ to the reward at each time step $t'$. As with standard policy gradients, a practical implementation of this method estimates the expectation by sampling trajectories from the current policy, and may be improved by following the natural gradient direction.

\subsection{Maximum Entropy Actor-Critic Algorithms}
\label{sec:maxentac}

Instead of directly differentiating the variational lower bound, we can adopt a message passing approach which, as we will see later, can produce lower-variance gradient estimates. First, note that we can write down the following equation for the optimal target distribution for $q(\ba_t|\bs_t)$:
\begin{equation*}
q^\star(\ba_t|\bs_t) = \frac{1}{Z}\exp \left(
E_{q(\bs_{(t+1):T},\ba_{(t+1):T}|\bs_t,\ba_t)} \left[ \sum_{t'=t}^T \log p(\op_{t'} | \bs_{t'},\ba_{t'}) - \sum_{t'=t+1}^T \log q(\ba_{t'}|\bs_{t'}) \right]
\right).
\end{equation*}
This is because conditioning on $\bs_t$ makes the action $\ba_t$ completely independent of all past states, but the action still depends on all future states and actions. Note that the dynamics terms $p(\bs_{t+1}|\bs_t,\ba_t)$ and $q(\bs_{t+1}|\bs_t,\ba_t)$ do not appear in the above equation, since they perfectly cancel. We can simplify the expectation above as follows:
\begin{align*}
&E_{q(\bs_{(t+1):T},\ba_{(t+1):T}|\bs_t,\ba_t)}[ \log p(\op_{t:T}|\bs_{t:T},\ba_{t:T}) ] = \\
&\log p(\op_t | \bs_t,\ba_t) +  E_{q(\bs_{t+1} | \bs_t, \ba_t)}\left[ E \left[ \sum_{t'=t+1}^T \log p(\op_{t'} | \bs_{t'},\ba_{t'}) - \log q(\ba_{t'}|\bs_{t'}) \right] \right].
\end{align*}
In this case, note that the inner expectation does not contain $\bs_t$ or $\ba_t$, and therefore makes for a natural representation for a message that can be sent from future states. We will denote this message $V(\bs_{t+1})$, since it will correspond to a soft value function:
\begin{align*}
V(\bs_t) &= E \left[ \sum_{t'=t+1}^T \log p(\op_{t'} | \bs_{t'},\ba_{t'}) - \log q(\ba_{t'}|\bs_{t'}) \right] \\
&= E_{q(\ba_t | \bs_t)}[\log p(\op_t | \bs_t,\ba_t) - \log q(\ba_t|\bs_t) + E_{q(\bs_{t+1}|\bs_t,\ba_t}[V(\bs_{t+1})] ].
\end{align*}
For convenience, we can also define a Q-function as
\begin{align*}
Q(\bs_t,\ba_t) = \log p(\op_t | \bs_t,\ba_t) + E_{q(\bs_{t+1}|\bs_t,\ba_t)}[V(\bs_{t+1})],
\end{align*}
such that $V(\bs_t) = E_{q(\ba_t | \bs_t)}[Q(\bs_t,\ba_t) - \log q(\ba_t|\bs_t)]$, and the optimal policy is
\begin{equation}
q^\star(\ba_t|\bs_t) = \frac{\exp(Q(\bs_t,\ba_t))}{\log \int_\mathcal{A} \exp(Q(\bs_t,\ba_t)) d\ba_t }. \label{eq:optpolicy}
\end{equation}
Note that, in this case, the value function and Q-function correspond to the values of the current policy $q(\ba_t|\bs_t)$, rather than the optimal value function and Q-function, as in the case of dynamic programming. However, at convergence, when $q(\ba_t|\bs_t) = q^\star(\ba_t|\bs_t)$ for each $t$, we have
\begin{align}
V(\bs_t) &= E_{q(\ba_t | \bs_t)}[Q(\bs_t,\ba_t) - \log q(\ba_t | \bs_t)] \nonumber \\
&= E_{q(\ba_t | \bs_t)}\left[ Q(\bs_t,\ba_t) - Q(\bs_t,\ba_t) + \log \int_\mathcal{A} \exp(Q(\bs_t,\ba_t)) d\ba_t \right] \nonumber \\
&= \log \int_\mathcal{A} \exp(Q(\bs_t,\ba_t)) d\ba_t, \label{eq:valuefunction}
\end{align}
which is the familiar soft maximum from Section~\ref{sec:maxentinference}. We now see that the optimal variational distribution for $q(\ba_t|\bs_t)$ can be computed by passing messages backward through time, and the messages are given by $V(\bs_t)$ and $Q(\bs_t,\ba_t)$.

So far, this derivation assumes that the policy and messages can be represented exactly. We can relax the first assumption in the same way as in the preceding section. We first write down the variational lower bound for a single factor $q(\ba_t|\bs_t)$ as following:
\begin{align}
\max_{q(\ba_t|\bs_t)} E_{\bs_t \sim q(\bs_t)} \left[ E_{\ba_t \sim q(\ba_t|\bs_t)}[ Q(\bs_t,\ba_t) - \log q(\ba_t|\bs_t) ] \right].\label{eq:variationalpolicyobj}
\end{align}
It's straightforward to show that this objective is simply the full variational lower bound, which is given by $E_{q(\tau)}[\log p(\tau) - \log q(\tau)]$, restricted to just the terms that include $q(\ba_t|\bs_t)$. If we restrict the class of policies $q(\ba_t|\bs_t)$ so that they cannot represent $q^\star(\ba_t|\bs_t)$ exactly, we can still optimize the objective in Equation~(\ref{eq:variationalpolicyobj}) by computing its gradient, which is given by
\begin{equation*}
E_{\bs_t \sim q(\bs_t)} \left[ E_{\ba_t \sim q(\ba_t|\bs_t)}[ \nabla \log q(\ba_t|\bs_t) (Q(\bs_t,\ba_t) - \log q(\ba_t|\bs_t) - b(\bs_t)) ] \right],
\end{equation*}
where $b(\bs_t)$ is any state-dependent baseline. This gradient can computed using samples from $q(\tau)$ and, like the policy gradient in the previous section, is directly analogous to a classic likelihood ratio policy gradient. The modification lies in the use of the backward message $Q(\bs_t,\ba_t)$ in place of the Monte Carlo advantage estimate. The algorithm therefore corresponds to an actor-critic algorithm, which generally provides lower variance gradient estimates.

In order to turn this into a practical algorithm, we must also be able to approximate the backward message $Q(\bs_t,\ba_t)$ and $V(\bs_t)$. A simple and straightforward approach is to represent them with parameterized functions $Q_\phi(\bs_t,\ba_t)$ and $V_\psi(\bs_t)$, with parameters $\phi$ and $\psi$, and optimize the parameters to minimize a squared error objectives
\begin{align}
\mathcal{E}(\phi) &= E_{(\bs_t,\ba_t) \sim q(\bs_t,\ba_t)}\left[ \left( r(\bs_t,\ba_t) + E_{q(\bs_{t+1}|\bs_t,\ba_t)}[V_\psi(\bs_{t+1})] - Q_\phi(\bs_t,\ba_t) \right)^2 \right] \label{eq:qerror} \\
\mathcal{E}(\psi) &= E_{\bs_t \sim q(\bs_t)}\left[ \left(
E_{\ba_t \sim q(\ba_t|\bs_t)} [ Q_\phi(\bs_t,\ba_t) - \log q(\ba_t|\bs_t) ]
- V_\psi(\bs_t,\ba_t)
\right)^2 \right]. \nonumber
\end{align}
This interpretation gives rise to a few interesting possibilities for maximum entropy actor-critic and policy iteration algorithms. First, it suggests that it may be beneficial to keep track of both $V(\bs_t)$ and $Q(\bs_t,\ba_t)$ networks. This is perfectly reasonable in a message passing framework, and in practice might have many of the same benefits as the use of a target network, where the updates to $Q$ and $V$ can be staggered or damped for stability. Second, it suggests that policy iteration or actor-critic methods might be preferred (over, for example, direct Q-learning), since they explicitly handle both approximate messages and approximate factors in the structured variational approximation. This is precisely the scheme employed by the soft actor-critic algorithm~\citep{sac}.

\subsection{Soft Q-Learning}
\label{sec:softqlearning}

We can derive an alternative form for a reinforcement learning algorithm without using an explicit policy parameterization, fitting only the messages $Q_\phi(\bs_t,\ba_t)$. In this case, we assume an implicit parameterization for both the value function $V(\bs_t)$ and policy $q(\ba_t | \bs_t)$, where
\[
V(\bs_t) = \log \int_\mathcal{A} \exp(Q(\bs_t,\ba_t)) d\ba_t ,
\]
as in Equation~(\ref{eq:valuefunction}), and
\[
q(\ba_t|\bs_t) = \exp(Q(\bs_t,\ba_t) - V(\bs_t)),
\]
which corresponds directly to Equation~(\ref{eq:optpolicy}). In this case, no further parameterization is needed beyond $Q_\phi(\bs_t,\ba_t)$, which can be learned by minimizing the error in Equation~(\ref{eq:qerror}), substituting the implicit equation for $V(\bs_t)$ in place of $V_\psi(\bs_t)$. We can write the resulting gradient update as
\[
\phi \leftarrow \phi - \alpha E\left[
\frac{d Q_\phi}{d\phi}(\bs_t,\ba_t) \left(
Q_\phi(\bs_t,\ba_t) - \left(
r(\bs_t,\ba_t) + \log \int_\mathcal{A} \exp(Q(\bs_{t+1},\ba_{t+1})) d\ba_{t+1}
\right)
\right)
\right].
\]
It is worth pointing out the similarity to the standard Q-learning update:
\[
\phi \leftarrow \phi - \alpha E\left[
\frac{d Q_\phi}{d\phi}(\bs_t,\ba_t) \left(
Q_\phi(\bs_t,\ba_t) - \left(
r(\bs_t,\ba_t) + \max_{\ba_{t+1}} Q_\phi(\bs_{t+1},\ba_{t+1}))
\right)
\right)
\right].
\]
Where the standard Q-learning update has a $\max$ over $\ba_{t+1}$, the soft Q-learning update has a ``soft'' max. As the magnitude of the reward increases, the soft update comes to resemble the hard update.

In the case of discrete actions, this update is straightforward to implement, since the integral is replaced with a summation, and the policy can be extracted simply by normalizing the Q-function. In the case of continuous actions, a further level of approximation is needed to evaluate the integral using samples. Sampling from the implicit policy is also non-trivial, and requires an approximate inference procedure, as discussed by Haarnoja et al.~\citep{haarnoja}.

We can further use this framework to illustrate an interesting connection between soft Q-learning and policy gradients. According to the definition of the policy in Equation~(\ref{eq:optpolicy}), which is defined entirely in terms of $Q_\phi(\bs_t,\ba_t)$, we can derive an alternative gradient with respect to $\phi$ starting from the policy gradient. This derivation represents a connection between policy gradient and Q-learning that is not apparent in the standard framework, but becomes apparent in the maximum entropy framework. The full derivation is provided by Haarnoja et al.~\citep{haarnoja} (Appendix B). The final gradient corresponds to
\[
\nabla_\phi J(\phi) = \sum_{t=1}^T E_{(\bs_t,\ba_t)\sim q(\bs_t,\ba_t)}\left[
\left( \nabla_\phi Q(\bs_t,\ba_t) - \nabla_\phi V(\bs_t) \right) \hat{A}(\bs_t,\ba_t)
\right].
\]
The soft Q-learning gradient can equivalently be written as
\begin{equation}
\nabla_\phi J(\phi) = \sum_{t=1}^T E_{(\bs_t,\ba_t)\sim q(\bs_t,\ba_t)}\left[
\nabla_\phi Q(\bs_t,\ba_t) \hat{A}(\bs_t,\ba_t)
\right], \label{eq:sqlgrad}
\end{equation}
where we substitute the target value $r(\bs_t,\ba_t) + V(\bs_{t+1})$ for $\hat{A}(\bs_t,\ba_t)$, taking advantage of the fact that we can use any state-dependent baseline. Although these gradients are not exactly equal, the extra term $- \nabla_\phi V(\bs_t)$ simply accounts for the fact that the policy gradient alone is insufficient to resolve one extra degree of freedom in $Q(\bs_t,\ba_t)$: the addition or subtraction of an action-independent constant. We can eliminate this term if we add the policy gradient with respect to $\phi$ together with Bellman error minimization for $V(\bs_t)$, which has the gradient
\[
\nabla_\phi V(\bs_t) E_{\ba_t \sim q(\ba_t|\bs_t)} \left[
r(\bs_t,\ba_t) + E_{\bs_{t+1} \sim q(\bs_{t+1}|\bs_t,\ba_t)}[V(\bs_{t+1})]
\right] = \nabla_\phi V(\bs_t) E_{\ba_t \sim q(\ba_t|\bs_t)}[\hat{Q}(\bs_t,\ba_t)].
\]
Noting that $\hat{Q}(\bs_t,\ba_t)$ is simply a (non-baselined) return estimate, we can show that the sum of the policy gradient and value gradient exactly matches Equation~(\ref{eq:sqlgrad}) for a particular choice of state-dependent baseline, since the term $\nabla_\phi V(\bs_t) E_{\ba_t \sim q(\ba_t|\bs_t)}[\hat{Q}(\bs_t,\ba_t)]$ cancels against the term $-\nabla_\phi V(\bs_t) \hat{A}(\bs_t,\ba_t)$ in expectation over $\ba_t$ when $\hat{A}(\bs_t,\ba_t) = \hat{Q}(\bs_t,\ba_t)$ (that is, when we use a baseline of zero). This completes the proof of a general equivalence between soft Q-learning and policy gradients.

\section{Review of Prior Work}
\label{sec:related}

In this section, we will discuss a variety of prior works that have sought to explore the connection between inference and control, make use of this connection to devise more effective learning algorithms, and extend it into other applications, such as intent inference and human behavior forecasting. We will first discuss the variety of frameworks proposed in prior work that are either equivalent to the approach presented in this article, or special cases (or generalization) thereof (Section~\ref{sec:prior_frameworks}). We will then discuss alternative formulations that, though similar, differ in some critical way (Section~\ref{sec:distinct}). We will then discuss specific reinforcement learning algorithms that build on the maximum entropy framework (Section~\ref{sec:algorithms}), and conclude with a discussion of applications of the maximum entropy framework in other areas, such as intent inference, human behavior modeling, and forecasting (Section~\ref{sec:modeling}).

\subsection{Frameworks for Control as Inference}
\label{sec:prior_frameworks}

Framing control, decision making, and reinforcement learning as a probabilistic inference and learning problem has a long history, going back to original work by Rudolf Kalman~\citep{kalman_filter}, who described how the Kalman smoothing algorithm can also be used to solve control problems with linear dynamics and quadratic costs (the ``linear-quadratic regulator'' or LQR setting). It is worth noting that, in the case of linear-quadratic systems, the maximum entropy solution is a linear-Gaussian policy where the mean corresponds exactly to the optimal \emph{deterministic} policy. This sometimes referred to as the Kalman duality~\citep{todorov_kalman_duality}. Unfortunately, this elegant duality does not in general hold for non-LQR systems: the maximum entropy framework for control as inference generalizes standard optimal control, and the optimal stochastic policy for a non-zero temperature (see Section~\ref{sec:alt_model}) does not in general have the optimal deterministic policy as its mean.

Subsequent work expanded further on the connection between control and inference. \citet{attias} proposed to implement a planning algorithm by means of the Baum-Welch-like method in HMM-style models. \citet{todorov_lmdp} formulated a class of reinforcement learning problems termed ``linearly-solvable'' MDPs (LMDPs). LMDPs correspond to the graphical model described in Section~\ref{sec:graphical_model}, but immediately marginalize out the actions or, equivalently, posit that actions are equivalent to the next state, allowing the entire framework to operate entirely on states rather than actions. This gives rise to a simple and elegant framework that is especially tractable in the tabular setting. In the domain of optimal control, \citet{kappen_oc} formulated a class of path integral control problems that also correspond to the graphical model discussed in Section~\ref{sec:graphical_model}, but derived starting from a continuous time formulation and formulated as a diffusion process. This continuous time generalization arrives at the same solution in discrete time, but requires considerably more stochastic processes machinery, so is not discussed in detail in this article. Similar work by Toussaint and colleagues~\citep{toussaint_soc,rawlik_soc} formulated graphical models for solving decision making problems, indirectly arriving at the same framework as in the previously discussed works. In the case of~\citet{toussaint_soc}, expectation propagation~\citep{ep} was adapted for approximate message passing during planning. \citet{ziebart_thesis} formulated learning in PGMs corresponding to the control or reinforcement learning problem as a problem of learning reward functions, and used this connection to derive maximum entropy \emph{inverse} reinforcement learning algorithms, which are discussed in more detail in Section~\ref{sec:modeling}. Ziebart's derivation of the relationship between decision making, conditional random fields, and PGMs also provides a thorough exploration of the foundational theory in this field, and is a highly recommend compendium to accompany this article for readers seeking a more in-depth theoretical discussion and connections to maximum entropy models~\citep{ziebart_thesis}. The particular PGM studied by Ziebart is discussed in Section~\ref{sec:alt_model}: although Ziebart frames the model as a conditional random field, without the auxiliary optimality variables $\opt$, this formulation is equivalent to the one discussed here. Furthermore, the maximum causal entropy method discussed by Ziebart can be shown to be equivalent to the variational inference formulation presented in Section~\ref{sec:fixeddynamics}.

\subsection{Related but Distinct Approaches}
\label{sec:distinct}

All of the methods discussed in the previous section are either special cases or generalizations of the control as inference framework presented in this article. A number of other works have presented related approaches that also aim to unify control and inference, but do so in somewhat different ways. We survey some of these prior techniques in this section, and describe their technical and practical differences from the presented formulation.

\paragraph{Boltzmann Exploration.} The form of the optimal policy in the maximum entropy framework (e.g., Equation~(\ref{eq:optpolicy})) suggests a very natural exploration strategy: actions that have large Q-value should be taken more often, while actions that have low Q-value should be taken less often, and the stochastic exploration strategy has the form of a Boltzmann-like distribution, with the Q-function acting as the negative energy. A large number of prior methods~\citep{suttonadp,KLMSurvey} have proposed to use such a policy distribution as an exploration strategy, but in the context of a reinforcement learning algorithm where the Q-function is learned via the standard (``hard'') max operator, corresponding to a temperature (see Section~\ref{sec:alt_model}) of zero. Boltzmann exploration therefore does not optimize the maximum entropy objective, but rather serves as a heuristic modification to enable improved exploration. A closely related idea is presented in the work on energy-based reinforcement learning~\citep{sallans,ebrl}, where the free energy of an energy-based model (in that case, a restricted Boltzmann machine) is adjusted based on a reinforcement learning update rule, such that the energy corresponds to the negative Q-function. Interestingly, energy-based reinforcement learning can optimize either the maximum entropy objective or the standard objective (with Boltzmann exploration), based on the type of update rule that is used. When used with an on-policy SARSA update rule, as proposed by \citet{sallans}, the method actually does optimize the maximum entropy objective, since the policy uses the Boltzmann distribution. However, when updated using an off-policy Q-learning objective with a hard max, the method reduces to Boltzmann exploration and optimizes the standard RL objective.

\paragraph{Entropy Regularization.} In the context of policy gradient and actor-critic methods, a commonly used technique is to use ``entropy regularization,'' where an entropy maximization term is added to the policy objective to prevent the policy from becoming too deterministic prematurely. This technique was proposed as early as the first work on the REINFORCE algorithm~\citep{WilliamsPeng91,williams}, and is often used in recent methods (see, e.g., discussion by~\citet{pgq}). While the particular technique for incorporating this entropy regularizer varies, typically the simplest way is to simply add the gradient of the policy entropy at each sampled state to a standard policy gradient estimate, which itself may use a critic. Note that this is \emph{not}, in general, equivalent to the maximum entropy objective, which not only optimizes for a policy with maximum entropy, but also optimizes the policy itself to \emph{visit} states where it has high entropy. Put another way, the maximum entropy objective optimizes the expectation of the entropy with respect to the policy's state distribution, while entropy regularization only optimizes the policy entropy at the states that are visited, without actually trying to modify the policy itself to visit high-entropy states (see, e.g., Equation (2) in \citet{pgq}). While this does correspond to a well-defined objective, that objective is rather involved to write out and generally not mentioned in work that uses entropy regularization. The technique is typically presented as a heuristic modification to the policy gradient. Interestingly, it is actually easier to perform proper maximum entropy RL than entropy regularization: maximum entropy RL with a policy gradient or actor-critic method only requires subtracting $\log \pi(\ba|\bs)$ from the reward function, while heuristic entropy maximization typically uses an explicit entropy gradient.

\paragraph{Variational Policy Search and Expectation Maximization.} Another formulation of the reinforcement learning problem with strong connections to probabilistic inference is the formulation of policy search in an expectation-maximization style algorithm. One common way to accomplish this is to directly treat rewards as a probability density, and then use a ``pseudo-likelihood'' written as
\[
J(\theta) = \int r(\tau) p(\tau | \theta) d\tau,
\]
where $r(\tau)$ is the total reward along a trajectory, and $p(\tau|\theta)$ is the probability of observing a trajectory $\tau$ given a policy parameter vector $\theta$. Assuming $r(\tau)$ is positive and bounded and applying Jensen's inequality results in a variety of algorithms~\citep{rwr,em_policy_search,neumann,mpo}, including reward-weighted regression~\citep{rwr}, that all follow the following general recipe: samples are weighted according to some function of their return (potentially with importance weights), and the policy is then updated by optimizing a regression objective to match the sample actions, weighted by these weights. The result is that samples with higher return are matched more closely, while those with low return are ignored. Variational policy search methods also fall into this category~\citep{neumann,levine_vgps}, sometimes with the modification of using explicit trajectory optimization rather than reweighting to construct the target actions~\citep{levine_vgps}, and sometimes using an exponential transformation on $r(\tau)$ to ensure positivity. Unlike the approach discussed in this article, the use of the reward as a ``pseudo-likelihood'' does not correspond directly to a well-defined probabilistic model, though the application of Jensen's inequality can still be motivated simply from the standpoint of deriving a bound for the RL optimization problem. A more serious disadvantage of this class of methods is that, by regressing onto the reweighted samples, the method loses the ability to properly handle risk for stochastic dynamics and policies. Consider, for example, a setting where we aim to fit a unimodal policy for a stateless problem with a 1D action. If we have a high reward for the action $-1$ and $+1$, and a low reward for the action $0$, the optimal fit will still place all of the probability mass in the middle, at the action $0$, which is the worst possible option. Mathematically, this problem steps from the fact that supervised learning matches a target distribution by minimizing a KL-divergence of the form $D_\text{KL}(p_\text{tgt} \| p_\theta)$, where $p_\text{tgt}$ is the target distribution (e.g., the reward or exponentiated reward). RL instead minimizes a KL-divergence of the form $D_\text{KL}(p_\theta \| p_\text{tgt})$, which prioritizes finding a mode of the target distribution rather than matching its moments. This issue is discussed in more detail in Section 5.3.5 of~\citep{levine_thesis}. In general, the issue manifests itself as risk-seeking behavior, though distinct in nature from the risk-seeking behavior discussed in Section~\ref{sec:maxentobjective}. Note that \citet{toussaint06} also propose an expectation-maximization based algorithm for control as inference, but in a framework that does in fact yield maximum expected reward solutions, with a similar formulation to the one in this article.

\paragraph{KL-Divergence Constraints for Policy Search.} Policy search methods frequently employ a constraint between the new policy and the old policy at each iteration, in order to bound the change in the policy distribution and thereby ensure smooth, stable convergence. Since policies are distributions, a natural choice for the form of this constraint is a bound on the KL-divergence between the new policy and the old one~\citep{covariant,reps,mfgps,trpo,mpo}. When we write out the Lagrangian of the resulting optimization problem, we typically end up with a maximum entropy optimization problem similar to the one in Equation~(\ref{eq:maxcausalentpolicy}), where instead of taking the KL-divergence between the new policy and exponentiated reward, we instead have a KL-divergence between the new policy and the old one. This corresponds to a maximum entropy optimization where the reward is $r(\bs,\ba) + \lambda \log \bar{\pi}(\ba|\bs)$, where $\lambda$ is the Lagrange multiplier and $\bar{\pi}$ is the old policy, and the entropy term has a weight of $\lambda$. This is equivalent to a maximum entropy optimization where the entropy has a weight of one, and the reward is scaled by $\frac{1}{\lambda}$. Thus, although none of these methods actually aim to optimize the maximum entropy objective in the end, each step of the policy update involves solving a maximum entropy problem. A similar approach is proposed by \citet{rawlik_soc}, where a sequence of maximum entropy problems is solved in a Q-learning style framework to eventually arrive at the standard maximum reward solution.

\subsection{Reinforcement Learning Algorithms}
\label{sec:algorithms}

Maximum entropy reinforcement learning algorithms have been proposed in a range of frameworks and with a wide variety of assumptions. The path integral framework has been used to derive algorithms for both optimal control and planning~\citep{kappen_oc} and policy search via reinforcement learning~\citep{pi2}. The framework of linearly solvable MDPs has been used to derive policy search algorithms~\citep{lmdp_policy}, value function based algorithms~\citep{todorov_lmdp}, and inverse reinforcement learning algorithms~\citep{ldmp_irl}. More recently, entropy maximization has been used as a component in algorithms based on model-free policy search with importance sampling and its variants~\citep{gps,pcl,trust_pcl}, model-based algorithms based on the guided policy search framework~\citep{mfgps,cgps,endtoend}, and a variety of methods based on soft Q-learning~\citep{haarnoja,schulman} and soft actor-critic algorithms~\citep{sac,hausman}.

The particular reasons for the use of the control as inference framework differ between each of these algorithms. The motivation for linearly solvable MDPs is typically based on computationally tractable exact solutions for tabular settings~\citep{lmdp_policy}, which are enabled essentially by dispensing with the non-linear maximization operator in the standard RL framework. Although the maximum entropy dynamic programming equations are still not linear in terms of value functions and Q-functions, they are linear under an exponential transformation. The reason for this is quite natural: since these methods implement sum-product message passing, the only operations in the original probability space are summations and multiplications. However, for larger problems where tabular representations are impractical, these benefits are not apparent.

In the case of path consistency methods~\citep{pcl,trust_pcl}, the maximum entropy framework offers an appealing mechanism for off-policy learning. In the case of guided policy search, it provides a natural method for matching distributions between model-based local policies and a global policy that unifies the local policy into a single globally coherent strategy~\citep{cgps,mfgps,endtoend}. For the more recent model-free maximum entropy algorithms, such as soft Q-learning~\citep{haarnoja} and soft-actor critic~\citep{sac}, as well the work of \citet{hausman}, the benefits are improved stability and model-free RL performance, improved exploration, and the ability to pre-train policies for diverse and under-specified goals. For example, \citet{haarnoja} present a quadrupedal robot locomotion task where the reward depends only on the speed of the robot's motion, regardless of direction. In a standard RL framework, this results in a policy that runs in an arbitrary direction. Under the maximum entropy framework, the optimal policy runs in all directions with equal probability. This makes it well-suited for pretraining general-purpose policies that can then be finetuned for more narrowly tailored tasks. More recently, Haarnoja et al. also showed that maximum entropy policies can be composed simply by adding their Q-functions, resulting in a Q-function with bounded difference against the optimal Q-function for the corresponding composed reward~\citep{haarnoja_composable}.

Recently, a number of papers have explored how the control as inference or maximum entropy reinforcement learning framework can be extended to add additional latent variables to the model, such that the policy is given by $\pi(\ba|\bs,\mathbf{z})$, where $\mathbf{z}$ is a latent variable. In one class of methods~\citep{hausman,learning_to_explore}, these variables are held constant over the duration of the episode, providing for a time-correlated exploration signal that can enable a single policy to capture multiple skills and rapidly explore plausible behaviors for new tasks by searching in the space of values for $\mathbf{z}$. In another class of methods~\citep{haarnoja,realnvp}, the latent variable $\mathbf{z}$ is selected independently at each time step, and the policy $\pi(\ba|\bs,\mathbf{z})$ has some simple unimodal form (e.g., a Gaussian distribution) conditioned on $\mathbf{z}$, but a complex multimodal form when $\mathbf{z}$ is integrated out. This enables the policy to represent very complex mulitmodal distributions, which can be useful, for example, for capturing the true maximum entropy distribution for an underspecified reward function (e.g., run in all possible directions). It also makes it possible to learn a higher-level policy that uses $\mathbf{z}$ as its action space~\citep{realnvp}, effectively driving the lower level policy and using it as a distribution over skills. This leads to a natural probabilistic hierarchical reinforcement learning formulation.

\subsection{Modeling, Intent Inference, and Forecasting}
\label{sec:modeling}

Aside from devising more effective reinforcement learning and optimal control algorithms, maximum entropy reinforcement learning has also been used extensively in the inverse reinforcement learning setting, where the goal is to infer intent, acquire reward functions from data, and predict the behavior of agents (e.g., humans) in the world from observation. Indeed, the use of the term ``maximum entropy reinforcement learning'' in this article is based on the work of Ziebart and colleagues, who proposed the maximum entropy \emph{inverse} reinforcement learning algorithm~\citep{maxentirl} for inferring reward functions and modeling human behavior.

While maximum entropy reinforcement learning corresponds to inference in the graphical model over the variables $\bs_t$, $\ba_t$, and $\opt$, \emph{inverse} reinforcement learning corresponds to a learning problem, where the goal is to learn the CPD $p(\opt|\bs_t,\ba_t)$, given example sequences $\{\bs_{1:T,i},\ba_{1:T,i},\op_{1:T,i}\}$, where $\opt$ is always true, indicating that the data consists of demonstrations of optimal trajectories. As with all graphical model learning problems, inference takes place in the inner loop of an iterative learning procedure. Exact inference via dynamic programming results in an algorithm where, at each iteration, we solve for the optimal soft value function, compute the corresponding policy, and then use this policy to compute the gradient of the likelihood of the data with respect to the parameters of the CPD $p(\opt|\bs_t,\ba_t)$. For example, if we use a linear reward representation, such that $p(\opt|\bs_t,\ba_t) = \exp(\phi^T f(\bs_t,\ba_t))$, the learning problem can be expressed as
\[
\phi^\star = \arg\max_\phi \sum_i \sum_t \log p(\ba_{t,i} | \bs_{t,i}, \op_{1:T}, \phi),
\]
where computing $\log p(\ba_{t,i} | \bs_{t,i}, \op_{1:T}, \phi)$ and its gradient requires solving for the optimal policy under the current reward parameters $\phi$.

The same optimism issue discussed in Section~\ref{sec:graphical_model} occurs in the inverse reinforcement learning setting, where exact inference in the graphical model produces an ``optimistic'' policy that assumes some degree of control over the system dynamics. For this reason, Ziebart and colleagues proposed the maximum causal entropy framework for inverse reinforcement learning under stochastic dynamics~\citep{maxcausalent}. Although this framework is derived starting from a causal reformulation of the maximum entropy principle, the resulting algorithm is exactly identical to the variational inference algorithm presented in Section~\ref{sec:variational}, and the corresponding learning procedure corresponds to optimizing the variational lower bound with respect to the reward parameters $\phi$.

Subsequent work in inverse reinforcement learning has studied settings where the reward function has a more complex, non-linear representation~\citep{gpirl,maxentdeepirl}, and extensions to approximate inference via the Laplace approximation (under known dynamics)~\citep{localioc,legibility} and approximate reinforcement learning (under unknown dynamics)~\citep{gcl,airl}. Aside from inferring reward functions from demonstrations for the purpose of imitation learning, prior work has also sought to leverage the framework of maximum entropy inverse reinforcement learning for inferring the intent of humans, for applications such as robotic assistance~\citep{legibility}, brain-computer interfaces~\citep{Javdani}, and forecasting of human behavior~\citep{kitani1,kitani2}.

Recent work has also drawn connections between generative adversarial networks (GANs)~\citep{gan} and maximum entropy inverse reinforcement learning~\citep{gcl,finn_adversarial,airl,gail}. This connection is quite natural since, just like generative adversarial networks, the graphical model in the maximum entropy reinforcement learning framework is a generative model, in this case of trajectories. GANs avoid the need for explicit estimation of the partition function by noting that, given a model $\hat{p}(\mathbf{x})$ for some true distribution $p(\mathbf{x})$, the optimal classifier for discriminating whether a sample $\mathbf{x}$ came from the model or from the data corresponds to the odds ratio
\[
D(\mathbf{x}) = \frac{p(\mathbf{x})}{p(\mathbf{x}) + \hat{p}(\mathbf{x})}.
\]
Although $\hat{p}(\mathbf{x})$ is unknown, fitting this ``discriminator'' and using its gradients with respect to $\mathbf{x}$ to modify $p(\mathbf{x})$ allows for effective training of the generative model. In the inverse reinforcement learning setting, the discriminator takes the form of the reward function. The reward function is learned so as to maximize the reward of the demonstration data and minimize the reward of samples from the current policy, while the policy is updated via the maximum entropy objective to maximize the expectation of the reward and maximize entropy. As discussed by Finn et al., this process corresponds to a generative adversarial network over trajectories, and also corresponds exactly to maximum entropy inverse reinforcement learning~\citep{finn_adversarial}. A recent extension of this framework also provides for an effective inverse reinforcement learning algorithm in a model-free deep RL context, as well as a mechanism for recovering robust and transferable rewards in ambiguous settings~\citep{airl}. A simplification on this setup known as generative adversarial imitation learning (GAIL)~\citep{gail} dispenses with the goal of recovering reward functions, and simply aims to clone the demonstrated policy. In this setup, the algorithm learns the advantage function directly, rather than the reward, which corresponds roughly to an adversarial version of the OptV algorithm~\citep{ldmp_irl}.

A number of prior works have also sought to incorporate probabilistic inference into a model of biological decision making and control~\citep{solway,botvinick,bottoussaint,friston}. The particular frameworks employed in these approaches differ: the formulation proposed by \citet{friston} is similar to the maximum entropy approach outlined in this survey, and also employs the formalism of approximate variational inference. The formulation described by \citet{bottoussaint} does not use the exponential reward transformation, and corresponds more closely to the ``pseudo-likelihood'' formulation outlined in Section~\ref{sec:distinct}.

\section{Perspectives and Future Directions}
\label{sec:future}

In this article, we discussed how the maximization of a reward function in Markov decision process can be formulated as an inference problem in a particular graphical model, and how a set of update equations similar to the well-known value function dynamic programming solution can be recovered as the direct consequence of applying structured variational inference to this graphical model. The classical maximum expected reward formulation emerges as a limiting case of this framework, while the general case corresponds to a maximum entropy variant of reinforcement learning or optimal control, where the optimal policy not only aims to maximize the expected reward, but also aims to maintain high entropy.

The framework of maximum entropy reinforcement learning has already been employed in a range of contexts, as discussed in the previous section, from devising more effective and powerful forward reinforcement learning algorithms, to developing probabilistic algorithms for modeling and reasoning about observed goal-driven behavior. A particularly exciting recent development is the intersection of maximum entropy reinforcement learning and latent variable models, where the graphical model for control as inference is augmented with additional variables for modeling time-correlated stochasticity for exploration~\citep{hausman,learning_to_explore} or higher-level control through learned latent action spaces~\citep{haarnoja,realnvp}. The extensibility and compositionality of graphical models can likely be leveraged to produce more sophisticated reinforcement learning methods, and the framework of probabilistic inference can offer a powerful toolkit for deriving effective and convergent learning algorithms for the corresponding models.

Less explored in the recent literature is the connection between maximum entropy reinforcement learning and robust control. Although some work has hinted at this connection~\citep{ziebart_thesis}, the potential for maximum entropy reinforcement learning to produce policies that are robust to modeling errors and distributional shift has not been explored in detail. In principle, a policy that is trained to achieve high expected reward under the highest possible amount of injected noise (highest entropy) should be robust to unexpected perturbations at test time. Indeed, recent work in robotics has illustrated that policies trained with maximum entropy reinforcement learning methods (e.g., soft Q-learning) do indeed exhibit a very high degree of robustness~\citep{haarnoja_composable}. However, a detailed theoretical exploration of this phenomenon has so far been lacking, and it is likely that it can be applied more broadly to a range of challenging problems involving domain shift, unexpected perturbations, and model errors.

Finally, the relationship between probabilistic inference and control can shed some light on the design of reward functions and objectives in reinforcement learning. This is an often-neglected topic that has tremendous practical implications: reinforcement learning algorithms typically assume that the reward function is an extrinsic and unchanging signal that is provided as part of the problem definition. However, in practice, the design of the reward function requires considerable care, and the success of a reinforcement learning application is in large part determined by the ability of the user to design a suitable reward function. The control as inference framework suggests a probabilistic interpretation of rewards as log probability of some discrete event variable $\opt$, and exploring how this interpretation can lead to more interpretable, more effective, and easier to specify reward functions could lead to substantially more practical reinforcement learning methods in the future.

\section*{Acknowledgements}

I would like to thank Emanuel Todorov, Karol Hausman, Nicolas Heess, and Shixiang Gu for suggestions on the writing, presentation, and prior work, as well as Vitchyr Pong, Rowan McAllister, Tuomas Haarnoja, and Justin Fu for feedback on earlier drafts of this tutorial and all the students and post-docs in the UC Berkeley Robotic AI \& Learning Lab for helpful discussion.

\bibliographystyle{apalike}
\bibliography{references}

\end{document}

%% file: defs.tex
\newcommand{\bo}{\mathbf{o}}
\newcommand{\bs}{\mathbf{s}}
\newcommand{\ba}{\mathbf{a}}

\newcommand{\op}{\mathcal{O}}
\newcommand{\opt}{\op_t}
\newcommand{\kl}{D_\text{KL}}
\newcommand{\ent}{\mathcal{H}}